%% file: acl_latex.tex
\pdfoutput=1

\documentclass[11pt]{article}

\usepackage[final]{acl}

\usepackage{times}
\usepackage{latexsym}

\usepackage[T1]{fontenc}

\usepackage[utf8]{inputenc}

\usepackage{microtype}

\usepackage{inconsolata}

\usepackage{multirow}
\usepackage{subfig}
\usepackage{amsmath}
\usepackage{booktabs}
\usepackage{amssymb}
\usepackage{bbm}
\usepackage{graphicx}
\usepackage{float}
\usepackage{algorithmic}
\usepackage{algorithm}
\usepackage{mathrsfs}

\title{CortexDebate: Debating Sparsely and Equally for Multi-Agent Debate}

\author{
 \textbf{Yiliu Sun\textsuperscript{1}},
 \textbf{Zicheng Zhao\textsuperscript{1}},
 \textbf{Sheng Wan\textsuperscript{2*}},
 \textbf{Chen Gong\textsuperscript{3*}}
\\
 \textsuperscript{1}School of Computer Science and Engineering, Nanjing University of Science and \\Technology, Nanjing, China.
 \\
 \textsuperscript{2}College of Artificial Intelligence, Nanjing Agricultural University, Nanjing, China.
\\
\textsuperscript{3}School of Automation and Intelligent Sensing, Shanghai Jiao Tong University, \\Shanghai, China.
\\
 \small{
   \textbf{Correspondence:} \href{wansheng315@hotmail.com}{wansheng315@hotmail.com}; \href{chen.gong@sjtu.edu.cn}{chen.gong@sjtu.edu.cn}.
 }
}

\begin{document}
\maketitle
\begin{abstract}
    Nowadays, single Large Language Model (LLM) struggles with critical issues such as hallucination and inadequate reasoning abilities. To mitigate these issues, Multi-Agent Debate (MAD) has emerged as an effective strategy, where LLM agents engage in in-depth debates with others on tasks. However, existing MAD methods face two major issues: (a) \textit{too lengthy input contexts}, which causes LLM agents to get lost in plenty of input information and experiences performance drop; and (b) \textit{the overconfidence dilemma}, where self-assured LLM agents dominate the debate, leading to low debating effectiveness. To address these limitations, we propose a novel MAD method called ``CortexDebate''. Inspired by the human brain's tendency to establish a sparse and dynamically optimized network among cortical areas governed by white matter, CortexDebate constructs a sparse debating graph among LLM agents, where each LLM agent only debates with the ones that are helpful to it. To optimize the graph, we propose a module named McKinsey-based Debate Matter (MDM), which acts as an artificial analog to white matter. By integrating the McKinsey Trust Formula, a well-established measure of trustworthiness from sociology, MDM enables credible evaluations that guide graph optimization. The effectiveness of our CortexDebate has been well demonstrated by extensive experimental results across eight datasets from four task types.
\end{abstract}

\input{introduction}

\input{related_work}
\input{preliminaries}
\input{approach}
\input{experiments}
\input{conclusion}
\input{limitations}




\bibliography{custom}

\appendix

\input{appendix/appendix_dataset_detail}
\input{appendix/supplementary_results}
\input{appendix/additional_experiment}
\input{appendix/performance_investigation}
\input{appendix/evaluation_strategy}
\input{appendix/cortexdebate_prompt}
\input{appendix/appendix_baseline_introduction}
\input{appendix/algorithm}

\end{document}

%% file: introduction.tex
\section{Introduction}
\label{introduction}

Recently, inspired by human cooperation, many multi-agent interaction methods~\cite{wan2024fusechat, xu2023recomp, tu2023unlocking, hu2024routerbench} have been proposed to further improve the reasoning results of LLMs. These methods aim to address critical issues faced by single LLM agent, such as hallucination and poor reasoning ability. Among these methods, Multi-Agent Debate (MAD)~\citep{zhang2024exploring, duimproving} has emerged as one of the most promising strategies, as it can effectively improve the performance of LLM agents through the debating process among them.

\begin{figure*}[t]
\centering
  \includegraphics[width=\linewidth]{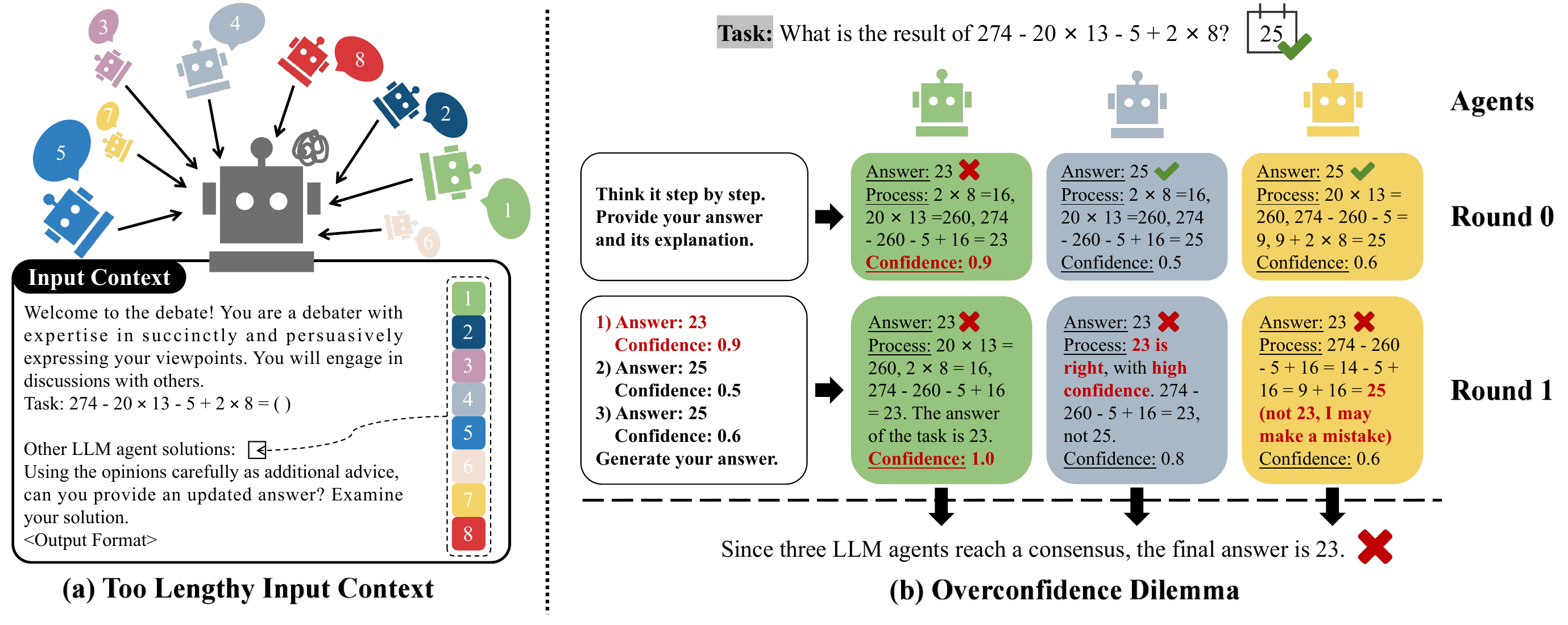}
  \caption {Shortcomings of the existing MAD methods: (a) Debating with all others causes lengthy contexts input to LLM agents. They may get lost in the vast amount of input information and perform unsatisfactorily; (b) Determining the debating impact of LLM agents simply based on their self-confidence may lead to the overconfident ones dominating the debate. This situation is harmful to the debating performance.}
  \label{figure1}
\end{figure*}

Although previous MAD methods have achieved promising results, they still suffer from two major shortcomings. As shown in Figure~\ref{figure1}, firstly, in these methods, each LLM agent is often required to debate with all other LLM agents, which causes its input context to expand significantly as the number of agents and debating rounds increase. Consequently, since single LLM agent usually struggles to handle lengthy input contexts~\cite {liu2024lost, luo2024graph}, it may get lost in the vast amount of input information, leading to a significant performance drop. Secondly, prior MAD methods determine the debating influence of each LLM agent simply according to its own confidence, which may lead to the overconfident LLM agents gradually dominating the entire debating process. As a result, the potential useful information provided by other ``weak’’ LLM agents may be ignored. Such unequal debate is harmful to debating effectiveness, as also confirmed by~\cite {xiong2023examining, xu2023towards}.

Therefore, inspired by the human cognition theory~\citep{thiebaut2022emergent}, this paper proposes a new MAD approach named CortexDebate, which mimics the working mode of the human brain cortex. As revealed by~\cite {thiebaut2022emergent}, given a problem, the human brain tends to establish a dynamic and sparse network among different cortical areas, and this network is gradually optimized by a specialized module named white matter. During the optimization process, the white matter focuses more on the influence between paired areas rather than the performance of a single cortical area.

By treating LLM agents as cortical areas in human brain, our proposed CortexDebate establishes a sparse and directed debating graph, where the nodes represent participated LLM agents and the edges carry information transmission between two LLM agents. Each directed graph edge is assigned a weight that reflects how much the performance of the tail LLM agent is expected to be improved by debating with the head LLM agent. Therefore, each tail LLM agent will not debate with the head LLM agents which do not help improve its performance. It means that the edges with small weights in the debating graph will be removed, resulting in a sparse graph. As a result, the length of context input to such tail LLM agent will also be reduced. To optimize the edge weights of the debating graph, akin to the white matter dynamically governing the optimization of sparse graph among different cortical areas in human brain, our CortexDebate introduces a module named McKinsey-based Debate Matter (MDM) that serves as the artificial white matter. To alleviate the overconfidence dilemma present in prior works, MDM considers both the performance of head LLM agent and the performance improvement expectation of tail LLM agent in deciding each edge weight. Specifically, MDM innovatively introduces McKinsey Trust Formula~\citep{lamarre2012mckinsey} to calculate edge weights, which has been widely used in sociology to evaluate the level of trustworthiness of a person through four aspects, including credibility, reliability, intimacy, and self-orientation. Among them, the first two evaluate individual abilities, while the last two evaluate the collaboration effectiveness with others. Therefore, this formula may suppress overconfident LLM agents, and also balance individual competence with teamwork ability of LLM agents in MAD.

The effectiveness of our CortexDebate has been well confirmed by the experiments on diverse tasks, including math, world knowledge question answering, reasoning, and long-context understanding. For instance, when compared with the state-of-the-art methods, in math task, CortexDebate increases \textit{Result Accuracy} (RA) by up to 9.00\% on GSM-IC dataset and 10.00\% on MATH dataset, respectively. In reasoning task, CortexDebate increases RA by up to 9.00\% on GPQA dataset and 12.33\% on ARC-C dataset, respectively. Besides, apart from achieving high performance, our CortexDebate significantly reduces the length of context input to each LLM agent, with a maximum reduction of 70.79\%.

The main contributions of this paper are summarized as follows:

1) We propose a new MAD method named CortexDebate, which can improve the performance of LLM agents by establishing a sparse and dynamic debating graph and reducing the burden of lengthy input context during the debate.

2) We propose a new module named MDM, which introduces McKinsey Trust Formula to evaluate both the confidence of each LLM agent and the usefulness to its debating component, thereby alleviating the overconfidence of LLM agents.

3) We conduct extensive experiments to show that our proposed CortexDebate outperforms representative baseline methods across multiple tasks such as math, world knowledge question answering, reasoning, and long-context understanding.

%% file: related_work.tex
\section{Related Work}
In a MAD system, each LLM agent presents its viewpoint and scrutinizes the viewpoints of other LLM agents across multiple rounds of debate~\citep{sun2024corex}. In summary, the existing MAD methods can be categorized as two types, namely \textit{sequential debate} and \textit{parallel debate}.

\textbf{Sequential Debate.} In these methods~\citep{hu2025debate, brownscalable, michael2023debate, wang2025learning, he2024agentscourt}, LLM agents generate their viewpoints in turn. Each LLM agent can only obtain the viewpoints of its preceding LLM agents. For example, \citet{liang2023encouraging} require two LLMs to refute each other in turn. In addition to debaters, \citet{guan2025mmd} add extra roles, such as judge and critic. The judge speaks before debaters to explain the task, and the critic speaks last to summarize debates. However, in a sequential debate system, each LLM agent must wait for previous LLM agents to finish reasoning before it starts. This makes debating time increase linearly with the number of LLM agents, leading to low efficiency which limits the scalability.

\textbf{Parallel Debate.} In these methods~\citep{phamlet, yin2023exchange, chern2024can, khandebating, liang2024debatrix, li2024coevol, hegazy2024diversity, zhang2024breaking}, all LLM agents simultaneously generate their viewpoints based on the viewpoints of other LLM agents in the last debating round. For example, \citet{chanchateval} require LLM agents to critique all answers generated in the last debating round and update its answer in each debating round simultaneously. In addition to the answers generated in the last round, \citet{sun2024towards} also provide each LLM agent with task-related information retrieved from the web. Besides, some methods~\citep{duan2024enhancing, yoffe2024debunc} try to adjust the debating influence of each LLM agent to improve the debating effectiveness. For example, \citet{chen2023reconcile} require each LLM agent to generate the confidence score for its own answer, and then inputs the score to other LLM agents along with the answer.

Since sequential debate systems face the low-efficiency issue mentioned above, our proposed CortexDebate follows the parallel debate framework. Compared with existing parallel debating methods which require each LLM agent to debate with all others in each round, our CortexDebate dynamically decides the necessary debating agents by establishing a sparse debating graph among all involved LLM agents, so that the input context to each agent can be shortened. This is also in contrary to~\citep{liu2024groupdebate, li2024improving} in which the debating opponents are fixed. Furthermore, different from prior methods which determine the debating impact of each LLM agent simply based on its own confidence, we introduce the McKinsey Trust Formula so that both the confidence of each LLM agent and the usefulness to its debating component can be evaluated.

%% file: preliminaries.tex
\section{Preliminaries}
\label{main_preliminaries}
In this section, we provide the problem definition for our CortexDebate, and introduce the McKinsey Trust Formula which plays an important role in our proposed CortexDebate.

\subsection{Problem Definition}
\label{problem_definition}

Our CortexDebate establishes a directed debating graph among $n$ LLM agents, $\mathcal{G}=(\mathcal{A}, \mathcal{E})$, where $\mathcal{A}=\left \{A_{i} \right \} _{i=1}^{n} $ is the vertex set representing participating LLM agents and $\mathcal{E}=\left \{E_{i\to j} \right \}_{i,j\in \left [1,2,...,n  \right ] }$ is the directed edge set representing information transmission. Here, each directed edge $E_{i\to j}$ is assigned a weight $W_{i\to j}$ that indicates the expected improvement in the performance of agent $A_{j}$ by debating with $A_{i}$. All the weights $\left \{ W_{i\to j}  \right \}$ are dynamically optimized during the debate process. Given a problem $Q$, the agents $\left \{ A_{i}  \right \}_{i=1}^{n}$ engage in $D$ rounds of debate. In the $d$-th debate round, each LLM agent $A_{i}$ scrutinizes the outputs of the LLM agents connected to it, and then generates its own output $O_{i}^{d}$ along with a self-confidence score $H_{i}^{d}$. Afterwards, the final answer of this debate round, $\textit{i.e.}$, $F_{d}$, is obtained by majority voting.

\begin{figure*}[t]
\centering
  \includegraphics[width=\linewidth]{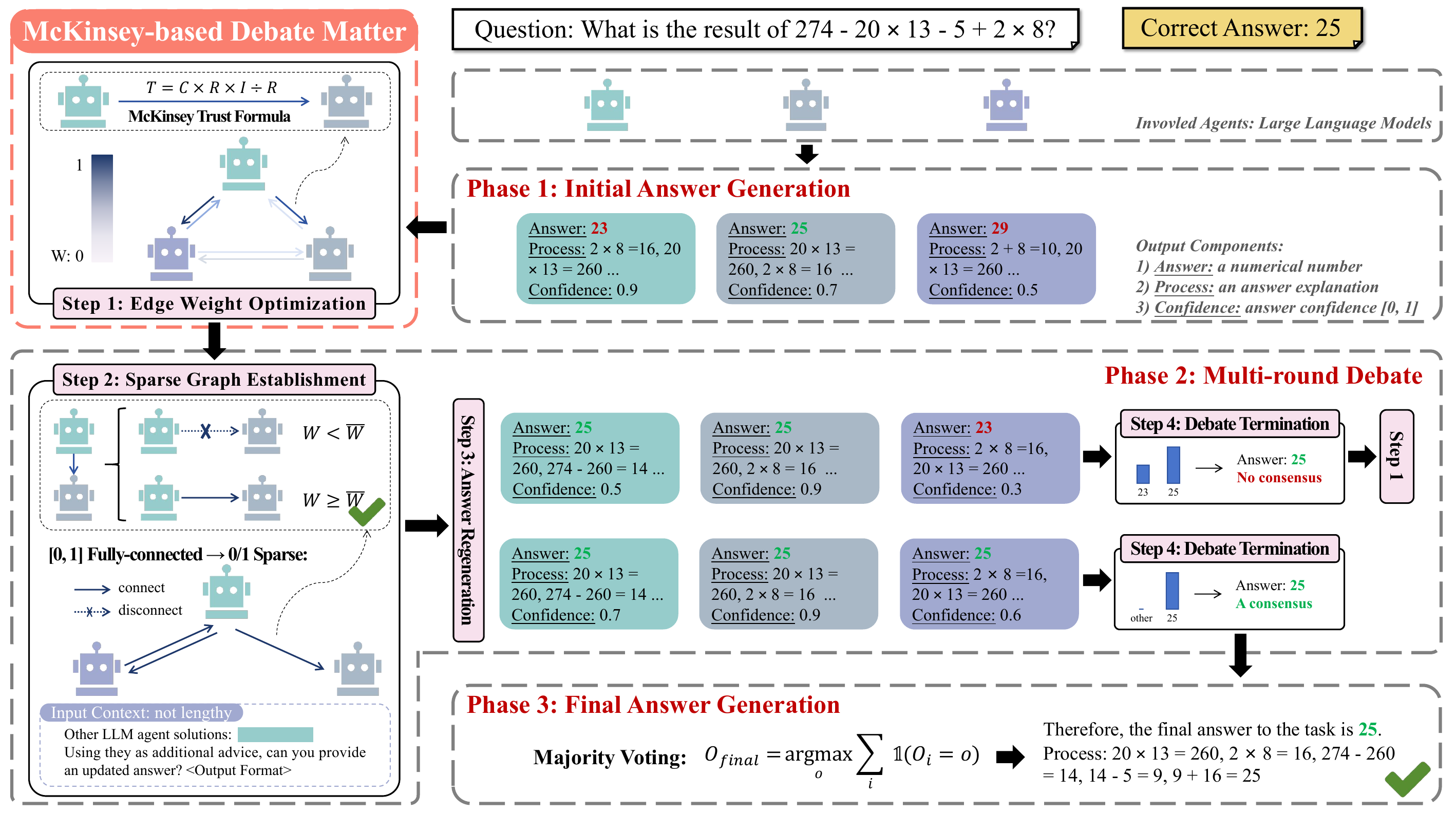}
  \caption {Overview of our proposed CortexDebate, which is inspired by the working mode of human brain cortex and consists of three phases: (a) Initial Answer Generation: Each LLM agent generates an answer, an explanation, and its confidence score. (b) Multi-round Debate: Participating LLM agents engage in debates guided by a sparse debating graph which is dynamically optimized by MDM module. (c) Final Answer Generation: After multi-round debates, the final answer is generated by majority voting.}
  \label{cortex}
\end{figure*}

\subsection{McKinsey Trust Formula}
\label{mckinsey_trust_formula}

The McKinsey Trust Formula~\citep{lamarre2012mckinsey} is widely used in sociology to evaluate the level of trustworthiness of a person within a group. This formula can be expressed as:
\begin{equation}
\setlength{\abovedisplayskip}{1mm}
\setlength{\belowdisplayskip}{1mm}
  \label{MTF}
  T= \frac{C\times R\times I}{S},
\end{equation}
\noindent where $C$, $R$, $I$, and $S$ denote credibility, reliability, intimacy, and self-orientation, respectively. Among them, credibility measures professional competence, reliability measures the stability of task performance, intimacy measures the relationship with the evaluated person, and self-orientation measures the self-orientation level of the evaluated person within a group.

In our MDM module, we adapt these four factors to the context of MAD. Specifically, for directed edge $E_{i\to j}$ connecting agent $A_{i}$ to $A_{j}$, credibility evaluates the professional competence of $A_{i}$. Reliability is the average confidence score of $A_{i}$ to its own answers in history debates, which represents the performance reliability on the current question. Intimacy represents the average degree of difference in viewpoints between $A_{i}$ and $A_{j}$ in history debates, as the collision of different viewpoints can enhance the debating effectiveness~\cite{xiong2023examining}. Self-orientation represents the participation level of $A_{i}$ in the debate (a lower participation level indicates higher self-orientation).

%% file: approach.tex
\section{Methodology}
\label{main_method}
In this section, we introduce the overall framework of our CortexDebate. As shown in Figure~\ref{cortex} and Algorithm~\ref{alg:1}, CortexDebate operates in three phases, including \textit{initial answer generation}, \textit{multi-round debate}, and \textit{final answer generation}. Unlike existing MAD methods that establish fully-connected and fixed graphs among LLM agents, our CortexDebate establishes a sparse and dynamic graph, where each LLM agent selectively debates with those that can contribute to its improvement. Besides, CortexDebate evaluates the performance of LLM agents and their usefulness to their debating components, enabling credible graph optimization.

\subsection{Phase 1: Initial Answer Generation}
\label{phase_1}
When given a problem $Q$, CortexDebate allows each LLM agent $A_{i}$ to independently generate an initial output $O_{i}^{0}$ and a self-confidence score $H_{i}^{0}$ (see Appendix~\ref{appendix_prompt_cortexdebate} for the specific prompt). To mitigate overconfidence, CortexDebate adopts a recalibration strategy, which has been proven to be effective in prior works~\cite{chen2023reconcile}. Our strategy can be expressed as:
\begin{equation}
\setlength{\abovedisplayskip}{1mm}
\setlength{\belowdisplayskip}{1mm}
  \label{recalibrate}
  H_{i}^{0} = \left\{\begin{matrix}0.8,
  &H_{i}^{0}\ge 0.8 \\0.6,
  &0.6\le H_{i}^{0}< 0.8 \\H_{i}^{0},
  &0.3\le H_{i}^{0}< 0.6 \\0.3,
  &H_{i}^{0}< 0.3
\end{matrix}\right..
\end{equation}

\subsection{Phase 2: Multi-round Debate}
\label{phase_2}
CortexDebate then comes into a debate phase, where the set of agents $\left \{A_{i} \right \}$ engage in $D$ rounds of debate. In the $d$-th debating round, CortexDebate comprises four steps, including \textit{edge weight optimization}, \textit{sparse graph establishment}, \textit{answer regeneration}, and \textit{debate termination}.

\textbf{Step 1: Edge Weight Optimization.} As the description of Equation~\eqref{MTF}, MDM calculates the edge weights based on four aspects, including \textit{credibility}, \textit{reliability}, \textit{intimacy}, and \textit{self-orientation}. Following the definition for each aspect in the context of MAD in Section~\ref{mckinsey_trust_formula}, the specific calculation of each aspect will be given next.

For $E_{i\to j}$, since the scaling law for LLM agents~\cite{hoffmann2022training} can evaluate abilities of one LLM agent, we use it to calculate credibility $C_{d}$, which can be expressed as:
\begin{equation}
\setlength{\abovedisplayskip}{1mm}
\setlength{\belowdisplayskip}{1mm}
\label{3}
  \mathcal{L}\left ( N,M \right ) = \frac{406.4}{N^{0.34} } +  \frac{410.7}{M^{0.28} }+  1.69, 
\end{equation}
\noindent where $N$, $M$, and $\mathcal{L}$ denote the parameter number, the token number of pre-training data, and the pre-training loss of one model, respectively. A smaller loss value indicates better model abilities, and thus $C_{d}$ is expressed as:
\begin{equation}
\setlength{\abovedisplayskip}{1mm}
\setlength{\belowdisplayskip}{1mm}
\label{4}
  C_{d} = \frac{1}{\mathcal{L}\left ( N,M \right )}. 
\end{equation}

For reliability $R_{d}$ which represents the average confidence score of $A_{i}$ in its own answers in the preceding $d-1$ rounds, its calculation can be expressed as:
\begin{equation}
\setlength{\abovedisplayskip}{1mm}
\setlength{\belowdisplayskip}{1mm}
\label{5}
  R_{d} = \frac{R_{d-1} \times (d-1)+  H_{i}^{d-1}}{d}.
\end{equation}

For intimacy $I_{d}$, which represents the average degree of difference in viewpoints between $A_{i}$ and $A_{j}$ in the preceding $d-1$ rounds, MDM first uses cosine similarity to calculate the textual similarity between $O_{i}^{d-1}$ and $O_{j}^{d-1}$. Subsequently, CortexDebate calculates the average viewpoint similarity between $A_{i}$ and $A_{j}$ in the preceding $d-1$ rounds, $\textit{i.e.}$, $\overline{{Sim}}_{d}$, as:
\begin{equation}
\setlength{\abovedisplayskip}{1mm}
\setlength{\belowdisplayskip}{1mm}
\label{6}
  \overline{{Sim}}_{d} \! \! = \!\! \frac{\overline{{Sim}}_{d-1} \!\! \times \!\! (d\!-\!1) \!\! + \!\! cos(O_{i}^{d-1} \! , \! O_{j}^{d-1})}{d},
\end{equation}
where $cos(a,b)$ calculates cosine similarity between $a$ and $b$. Since $I_{d}$ represents the average degree of difference, it is calculated as:
\begin{equation}
\setlength{\abovedisplayskip}{1mm}
\setlength{\belowdisplayskip}{1mm}
\label{7}
  I_{d} = 1-\overline{{Sim}}_{d}.
\end{equation}

For self-orientation $S_{d}$, based on the fact that less group participation indicates that one is more selfish, the MDM module uses the number of times that $A_{i}$ has debated with other LLM agents in the preceding $d-1$ rounds, denoted as $P_{d}$, to indirectly reflect self-orientation. The calculation can be expressed as:
\begin{equation}
\setlength{\abovedisplayskip}{1mm}
\setlength{\belowdisplayskip}{1mm}
\label{8}
  S_{d}=(d-1) \times (n-1) - P_{d},
\end{equation}
\noindent where $(d-1) \times (n-1)$ denotes the maximum number of times that one LLM agent can debate with others in the preceding $d-1$ rounds.

Therefore, following Equation~\eqref{MTF}, the weight of edge $E_{i\to j}$ can be calculated as:
\begin{equation}
\setlength{\abovedisplayskip}{1mm}
\setlength{\belowdisplayskip}{1mm}
\label{9}
  W_{i\to j}^{d}= \frac{C_{d}\times R_{d}\times I_{d}}{S_{d}}.
\end{equation}

\textbf{Step 2: Sparse Graph Establishment.} For $A_{j}$, it can debate with the other $n-1$ LLM agents. In other words, there are $n-1$ directed edges pointing to it, with $A_{j}$ as the tail node. CortexDebate determines the set of debating opponents for $A_{j}$ according to the weights of these edges $\left \{ W_{i\to j}^{d}  \right \}_{i=1, \ i \ne j}^{n}$.

Firstly, the average weight of these edges, $\textit{i.e.}$, $\overline{{W }}_{j}^{d}$, is calculated as:
\begin{equation}
\setlength{\abovedisplayskip}{1mm}
\setlength{\belowdisplayskip}{1mm}
\label{10}
  \overline{{W }}_{j}^{d} = \frac{1}{n-1} {\textstyle \sum_{i(i\ne j)}^{} W_{i\to j}^{d}}.
\end{equation}

Secondly, the edges with weights below $\overline{{W_{j}^{d} }}$ are removed, resulting in a sparse debating graph. The process can be expressed as:
\begin{equation}
\setlength{\abovedisplayskip}{1mm}
\setlength{\belowdisplayskip}{1mm}
\label{11}
  W_{i\to j}^{d} = \left\{\begin{matrix}1,
  & W_{i\to j}\ge  \overline{{W }}_{j}^{d} \\0,
  & W_{i\to j}<   \overline{{W }}_{j}^{d}
\end{matrix}\right..
\end{equation}
Therefore, the debating opponents for $A_{j}$, denoted as $Deb_{j}$, can be expressed as:
\begin{equation}
\setlength{\abovedisplayskip}{1mm}
\setlength{\belowdisplayskip}{1mm}
\label{12}
  Deb_{j}^{d} = \left \{ A_{i}\mid W_{i\to j}^{d}= 1, i\ne j \right \}.
\end{equation}

\textbf{Step 3: Answer Regeneration.} For LLM agent $A_{j}$, it receives the answers of the LLM agents in $Deb_{j}^{d}$, which are generated in the $(d-1)$-th debating round. Afterwards, $A_{j}$ needs to read and scrutinize these answers, and generate its new answer $O_{j}^{d}$ and self-confidence score $H_{j}^{d}$. The input prompt can be expressed as:
\begin{equation}
\setlength{\abovedisplayskip}{1mm}
\setlength{\belowdisplayskip}{1mm}
\label{13}
  Prompt_{j}^{d} = \left [ Ins,Q,\left \{ O_{k}^{d-1} \right \}  \right ],
\end{equation}
\noindent where $Ins$ denotes the instruction that stimulates $A_{j}$ to regenerate its answer and $\left \{ O_{k}^{d-1} \right \}$ denotes the set of answers that $A_{j}$ receives. The specific prompt is shown in Appendix~\ref{appendix_prompt_cortexdebate}.

\textbf{Step 4: Debate Termination.} After all the LLM agents have generated their answers, CortexDebate checks whether all the LLM agents reach a consensus (\textit{i.e.}, all the LLM agents agree on the same answer) or the debate reaches the maximum rounds. If so, the whole debating process concludes immediately.

\subsection{Phase 3: Final Answer Generation}

Once the entire debating process concludes, CortexDebate generates the final answer to the question by majority voting among all the answers generated in the last debating round, which can be expressed as:
\begin{equation}
\setlength{\abovedisplayskip}{1mm}
\setlength{\belowdisplayskip}{1mm}
  \label{14}
  O_{final} =   \mathop{\arg\max }\limits_{o} \sum_{i}^{} \mathbbm{1} \left ( O_{i} = o \right ),
\end{equation}
\noindent where $o$ denotes a distinct answer generated by any of the LLM agents. If all the generated answers are different after debates, we treat the final result as incorrect. By excluding fallback strategies, we can clearly attribute the observed performance solely to the debate mechanism itself.

%% file: experiments.tex
\section{Experiments}
This section introduces the experimental setup, experimental results, and analysis of our experiments. 

\input{table/main_result_acc}

\subsection{Experimental Setup}
\label{main_experimental_setup}
In this part, we introduce the details of the experimental setup.

\textbf{Tasks.} In our experiments, we consider four typical tasks, namely: (a) math task, (b) world knowledge question answering task, (c) reasoning task, and (d) long-context understanding task. For the math task, we use GSM-IC~\citep{shi2023large} and MATH~\citep{hendrycks2measuring} datasets. For the world knowledge question answering task, we use MMLU~\citep{hendrycksmeasuring} and MMLU-pro~\citep{wang2024mmlu} datasets. For the reasoning task, we use GPQA~\citep{rein2023gpqa} and ARC-C~\citep{clark2018think} dataset. For the long-context understanding task, we use LongBench~\citep{bai2023longbench} and SQuAD~\citep{rajpurkar2016squad} datasets. More details on the employed datasets for experiments can be found in Appendix~\ref{appendix_dataset_detail}.

\textbf{Evaluation Metrics.} For LongBench dataset, we follow~\cite{bai2023longbench} and utilize the \textit{Macro-Average} (M-Avg), which calculates the average score over major sub-task categories. For SQuAD dataset, we follow~\citep{rajpurkar2016squad} and utilize the \textit{Exact Match} (EM), which calculates the percentage of outputs containing correct answers. For the remaining six datasets, we follow~\citep{shi2023large, hendrycks2measuring, hendrycksmeasuring, wang2024mmlu, rein2023gpqa, clark2018think, sun2025fast} and utilize the \textit{Result Accuracy} (RA), which calculates the percentage of correct results.

\textbf{Baseline Methods.} Our proposed CortexDebate is compared with the three categories of methods: 1) \textbf{No debate:} Multi-agent Voting (MaV)~\citep{wangself}, 2) \textbf{Full debate:} Multi-LLM Debate (MLD)~\citep{duimproving}, RECONCILE~\citep{chen2023reconcile}, ChatEval~\citep{chanchateval}, and Peer Review Debate (PRD)~\citep{xu2023towards}, 3) \textbf{Part debate:} GroupDebate (GD)~\citep{liu2024groupdebate} and Neighbor Debate (ND)~\citep{li2024improving}. Among them, no debate methods are the multi-agent methods without using debating strategies, full debate methods are the MAD methods where each LLM agents are required to debate with all others, and part debate methods are the MAD methods where each LLM agents only debates with part of the others. Detailed introduction of these baseline methods can be found in Appendix~\ref{appendix_baseline_methods}.

For fairness, the maximum number of debating rounds is set to 5 for all debating methods.

\textbf{Backbone Models.} The backbone models involved in the debating system for our experiments are Qwen-2.5-7B-Instruct-Turbo~\cite{qwen2.5}, Mistral-7B-Instruct~\cite{jiang2023mistral}, Typhoon-1.5-8B-Instruct~\cite{pipatanakul2023typhoon}, Llama-3.1-8B-Instruct-Turbo~\cite{dubey2024llama}, and Gemma-2-9B-Instruct~\cite{yang2024weighted}. For simplicity, we refer to them as Qwen, Mistral, Typhoon, Llama, and Gemma, respectively.

\textbf{Implementation Details.} We follow prior works~\cite{duimproving, chen2023reconcile, besta2024graph} to experiment on a subset of 100 examples for each dataset. For each experiment, we conduct three runs on the same examples with the same setups and report average results along with their variances. We also conduct large-scale experiments on the more challenging datasets from each task (\textit{i.e.} MATH, MMLU-pro, GPQA, and LongBench) and observe similar results, which are detailed in Appendix~\ref{appendix_additional_experiments}.

\begin{figure*}[t]
\centering
  \includegraphics[width=\linewidth]{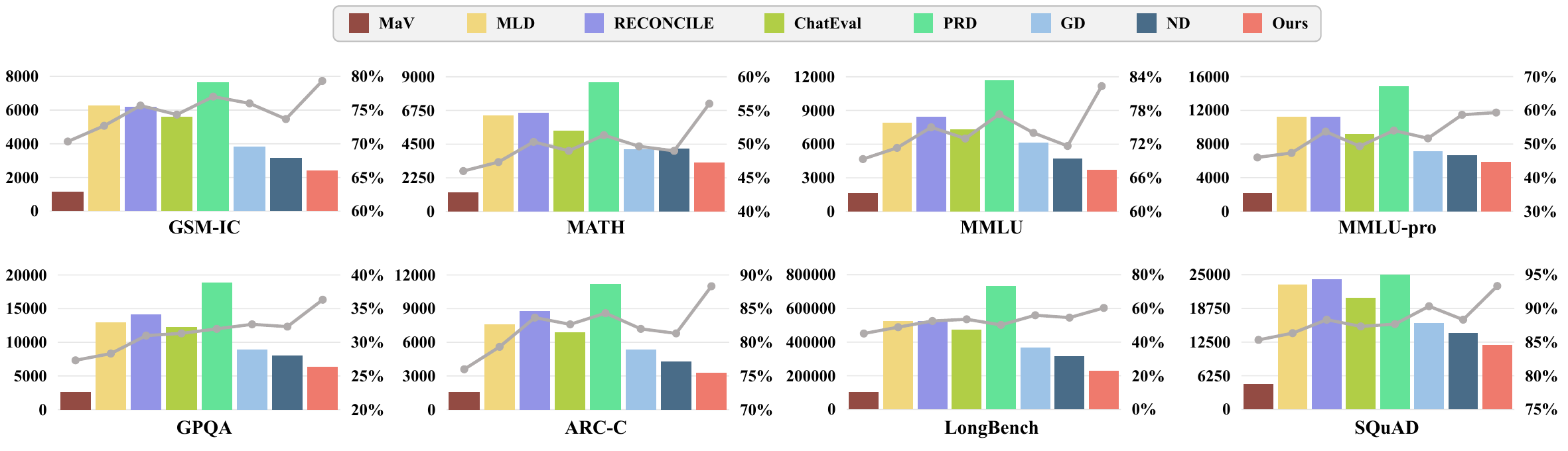}
  \caption {Comparison results of average length of context input to the LLM agents on eight datasets. We reflect the length of one input context through its token number. In each combined chart, the left vertical axis (representing token number) corresponds to the bar chart, while the right vertical axis (representing task accuracy) corresponds to the line chart.}
  \label{figure2}
\end{figure*}

\subsection{Main Results}
\label{main_results}

In this part, we present the experimental results and detailed analysis to highlight the effectiveness of our proposed CortexDebate.

\textbf{CortexDebate outperforms baseline methods.} Table~\ref{main_result_acc} reports the accuracy of our CortexDebate and baseline methods on eight datasets. Compared with the baseline methods, our CortexDebate achieves the highest accuracy and performs stably on all adopted datasets. Besides, we can find that the effectiveness and stability of the full debate methods (\textit{i.e.}, MLD, RECONCILE, ChatEval, and PRD) drops on complex reasoning and long-context tasks (\textit{i.e.}, GPQA, LongBench, and SQuAD). It is because the reasoning process increases with the complexity of the task, leading to the lengthy context issue mentioned in Section~\ref{introduction}. However, our CortexDebate still performs well and stably due to its sparse debating graph which reduces input context length and MDM module which makes each LLM agent debate with those that are helpful to it.

\textbf{CortexDebate significantly reduces input context length.} For each adopted dataset, we calculate the average token number of context input to a single LLM agent in each method and present the results in Figure~\ref{figure2}. Compared with MaV, MAD methods generally incur long context input to each LLM agent, indicating a significant challenge in reducing input context length while maintaining superior accuracy in MAD methods. Our proposed CortexDebate takes a further step, as it achieves both shorter input context length and higher task performance compared with other MAD baseline methods. The specific numerical values of the results shown in Figure~\ref{figure2} are presented in Appendix~\ref{appendix_additional_results}.

\begin{figure}[!t]
\centering
\subfloat[Average scores of our CortexDebate and baseline methods after each debating round.]{
		\includegraphics[scale=0.45]{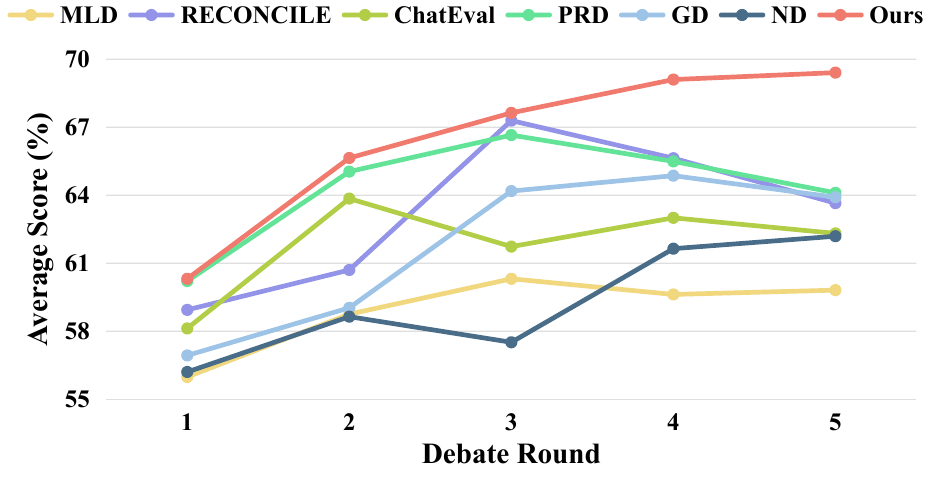}
        \label{fig_a}}\\
\subfloat[Proportion of examples achieving answer consensus.]{
		\includegraphics[scale=0.45]{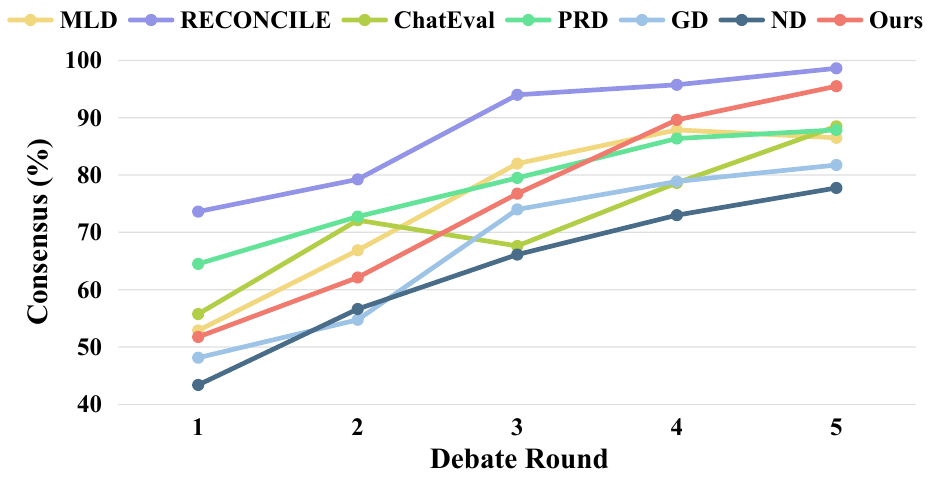}
        \label{fig_b}}
\caption{Results of average task scores and consensus proportions for MAD methods after each round.}
\label{round_accuracy_consensus}
\end{figure}

\textbf{CortexDebate debates effectively and equally.} Engaging in more effective debates is what MAD systems strive for. To study this, in Figures~\ref{fig_a} and~\ref{fig_b}, we plot the average scores and proportion of examples achieving consensus on the answers on eight adopted datasets after each debating round, respectively. From Figure~\ref{fig_a}, we have two important observations: (a) As the debate proceeds, the performance of our CortexDebate continues to improve. (b) Compared with the baseline methods, our CortexDebate maintains superior performance and achieves the highest score of 69.41\%. From Figure~\ref{fig_b}, our observations are likewise twofold: (a) In the initial rounds, since CortexDebate encourages the equal collision of different viewpoints, its consensus proportion is relatively low. However, as the debate proceeds, a high consensus proportion is achieved. (b) Compared with other methods, RECONCILE maintains the highest consensus proportion while its score fluctuates as shown in Figure~\ref{fig_a}. This is due to the overconfidence-caused unequal debate, where the debate is dominated by a few LLM agents and others tend to surrender. Differently, our CortexDebate alleviates this issue and maintains equally debates among LLM agents, thereby achieving consistent growth in score and consensus proportion. The numerical results are presented in Appendix~\ref{appendix_additional_results}.

\subsection{Performance Investigation}
\label{performance_investigation}

In this section, we conduct in-depth investigation on our CortexDebate to analyze its effectiveness. For each method, we use its average score on eight adopted datasets to represent its performance.


\input{table/each_part}

\input{table/cooperation}

\textbf{Each component of CortexDebate is indispensable.} To show that every component of CortexDebate (\textit{i.e.}, sparse debating graph and MDM module) is indispensable, we conduct an ablation study. For the fully-connected graph, we follow the basic MAD framework where each LLM agent debates with all others. For the sparse graph, we use different evaluation strategies to optimize the edge weights of the debating graph, including self-evaluation~\cite{chen2023reconcile}, peer evaluation~\cite{xu2023towards}, MDM (w/o $I_{d}$ and $S_{d}$ in Equation~\eqref{9}), and MDM (see Appendix~\ref{appendix_evaluation_strategies} for detailed introduction). As shown in Table~\ref{each_part}, compared with ``fully-connected graph + MDM'', ``sparse graph + MDM'' increases the average score by 5.65\%. It is because sparse debating graph structure alleviates lengthy input context issue and allows LLM agents to make full use of their input information. For different optimization strategies, the average task score of self-evaluation is only 62.13\%. It is due to the overconfidence dilemma mentioned in Section~\ref{introduction}. Peer evaluation and MDM (w/o $I_{d}$ and $S_{d}$ in Equation~\eqref{9}) alleviate this issue, achieving better performance compared with self-evaluation. Moreover, MDM further improves the task performance, since it considers both the performance of each LLM agent and the usefulness to its debating components, thereby conducting more credible evaluations compared with Peer evaluation and MDM (w/o $I_{d}$ and $S_{d}$ in Equation~\eqref{9}) which only evaluate individual performance.

\begin{figure}[t]
\centering
  \includegraphics[width=.9\linewidth]{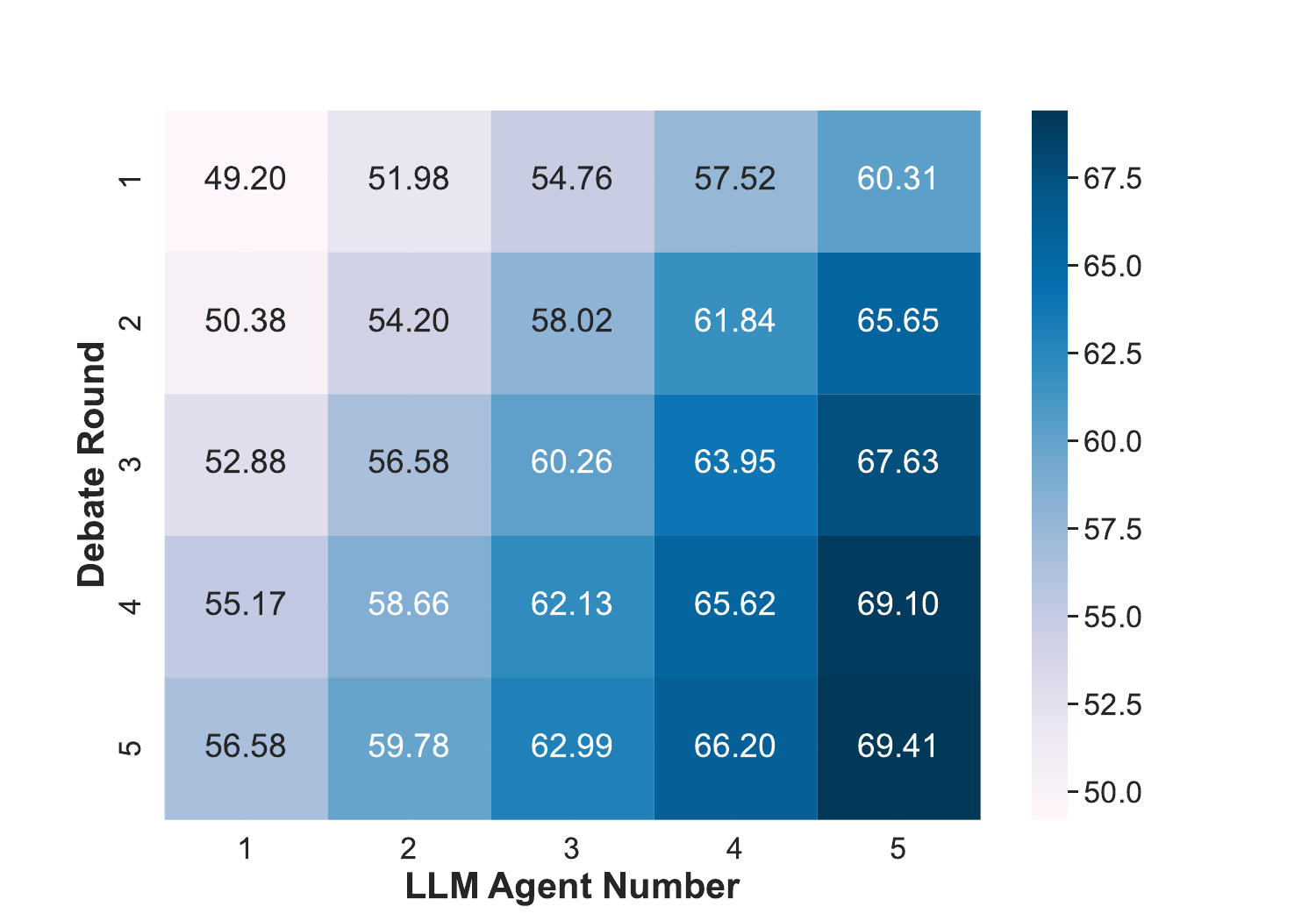}
  \caption {Task performance of CortexDebate under different LLM agent numbers and debating rounds.}
  \label{pic_sacle_number}
\end{figure}

\input{table/helpful}

\textbf{Considering cooperation performance among agents is essential.} On MATH dataset, we compare the average numbers of different viewpoint collisions (DVC) and correct viewpoint revision (CVR) per question of our CortexDebate with and without using intimacy (I) and self-orientation (S) factors. As shown in Table~\ref{cooperation}, we can find that incorporating I and S increases both the frequency and quality of viewpoint interactions. This confirms that considering interactions among agents is essential for enhancing collective reasoning in LLM-based systems.

\textbf{CortexDebate excels in large-scale debates.} To explore the influence of LLM agent number and debating rounds on our CortexDebate, we evaluate the task performance of CortexDebate under different numbers of participating LLM agents and debating rounds. We present the results in Figure~\ref{pic_sacle_number}. We can see that as the number of LLM agents and the debating rounds increase, the task performance of our CortexDebate continues to improve. Moreover, compared with debating rounds, the increase in the number of LLM agents contributes more to the performance improvement of CortexDebate. These results demonstrate the potential of CortexDebate for application in large-scale debates.

\textbf{CortexDebate retains helpful debates.} To further show the effectiveness of the retained edges and corresponding nodes in our CortexDebate, we conduct a detailed analysis. On MATH dataset, we calculate the average debating number (Avg-DN) and RA of every participated LLM agent under the fully-connected graph (FcG) and our proposed debating graph. As shown in Table~\ref{helpful}, our proposed CortexDebate reduces debates among the LLM agents, while improving the performance of every LLM agent. It suggests that our CortexDebate retains helpful debates while pruning harmful ones.

%% file: table/main_result_acc.tex
\begin{table*}
\small
  \centering
  \setlength{\tabcolsep}{1mm}
  \begin{tabular}{c|c|cccccc|cc}
\toprule
\multirow{2}{*}{Type}    & \multirow{2}{*}{Method}                      & GSM-IC               & MATH                 & MMLU                 & MMLU-pro             & GPQA                 & ARC-C        & LongBench            & SQuAD                \\ \cmidrule{3-10}
&               & \multicolumn{6}{c|}{RA $\uparrow$}    & M-Avg $\uparrow$    & EM $\uparrow$  \\ \midrule
No Debate            & MaV       & 70.33{\tiny $\pm$1.56}          & 46.00{\tiny $\pm$2.67}          & 69.33{\tiny $\pm$0.22}          & 46.00{\tiny $\pm$4.67}          & 27.33{\tiny $\pm$2.89}          & 76.00{\tiny $\pm$0.67}          & 45.11{\tiny $\pm$1.09}          & 85.33{\tiny $\pm$1.56}          \\ \midrule
\multirow{4}{*}{Full Debate}   & MLD        & 72.67{\tiny $\pm$0.22}          & 47.33{\tiny $\pm$0.89}          & 71.33{\tiny $\pm$1.56}          & 47.33{\tiny $\pm$0.89}          & 28.33{\tiny $\pm$2.89}          & 79.33{\tiny $\pm$0.22}          & 48.87{\tiny $\pm$2.21}          & 86.33{\tiny $\pm$0.22}          \\
& RECONCILE         & 75.67{\tiny $\pm$0.22}          & 50.33{\tiny $\pm$4.22}          & 75.00{\tiny $\pm$2.67}          & 53.67{\tiny $\pm$2.89}          & 31.00{\tiny $\pm$0.67}          & 83.67{\tiny $\pm$2.89}          & 52.55{\tiny $\pm$2.68}          & 88.33{\tiny $\pm$6.89}          \\
& ChatEval          & 74.33{\tiny $\pm$0.89}          & 49.00{\tiny $\pm$0.67}          & 73.00{\tiny $\pm$0.67}          & 49.33{\tiny $\pm$0.89}          & 31.33{\tiny $\pm$0.89}          & 82.67{\tiny $\pm$1.56}          & 53.56{\tiny $\pm$6.16}          & 87.33{\tiny $\pm$6.22}          \\
& PRD               & 77.00{\tiny $\pm$0.67}          & 51.33{\tiny $\pm$0.89}          & 77.33{\tiny $\pm$1.56}          & 54.00{\tiny $\pm$0.67}          & 32.00{\tiny $\pm$2.00}          & 84.33{\tiny $\pm$0.89}          & 50.21{\tiny $\pm$6.09}          & 87.67{\tiny $\pm$4.22}          \\  \midrule
\multirow{3}{*}{Part Debate}                       &GD                       & 76.00{\tiny $\pm$2.67}          & 49.67{\tiny $\pm$1.56}          & 74.00{\tiny $\pm$2.67}          & 51.67{\tiny $\pm$0.89}          & 32.67{\tiny $\pm$0.22}          & 82.00{\tiny $\pm$2.00}          & 55.97{\tiny $\pm$0.59}          & 90.33{\tiny $\pm$0.89}          \\
& ND     & 73.67{\tiny $\pm$1.56}          & 49.00{\tiny $\pm$0.67}          & 71.67{\tiny $\pm$2.89}          & 48.67{\tiny $\pm$1.56}          & 32.33{\tiny $\pm$1.56}          & 81.33{\tiny $\pm$2.89}          & 54.55{\tiny $\pm$6.18}          & 88.33{\tiny $\pm$1.56}          \\  \cmidrule{2-10}
& Ours   & \textbf{79.33{\tiny $\pm$0.22}} & \textbf{56.00{\tiny $\pm$0.67}} & \textbf{82.33{\tiny $\pm$0.22}} & \textbf{59.33{\tiny $\pm$0.22}} & \textbf{36.33{\tiny $\pm$1.56}} & \textbf{88.33{\tiny $\pm$0.89}} & \textbf{60.31{\tiny $\pm$0.32}} & \textbf{93.33{\tiny $\pm$0.89}} \\ \bottomrule
\end{tabular}
  \caption{\label{main_result_acc}
   Comparison results on the four different types of tasks. The unit of all the results is ``$\%$''. The format of the results is ``(average result)$\pm$(variance)''. ``$\uparrow$'' means that higher values are better. The best records under each metric are highlighted in bold.
  }
\end{table*}

%% file: table/each_part.tex
\begin{table}
  \centering
  \setlength{\tabcolsep}{0mm}
\begin{tabular}{lc}
\toprule
\multicolumn{1}{c}{Method}               & Score $\uparrow$ \\ \midrule
Fully-connected Graph     & 60.49         \\
+ MDM                & 63.76         \\ \midrule
Sparse Graph       & 62.72         \\
+ Self-evaluation (RECONCILE) & 62.13     \\
+ Peer Evaluation (PRD) & 66.71     \\
+ MDM (w/o $I_{d}$ and $S_{d}$ in Equation~\eqref{9}) & 66.69     \\
+ MDM (Ours) & \textbf{69.41}     \\ \bottomrule
\end{tabular}
  \caption{\label{each_part}
   Ablation study on our proposed CortexDebate. ``$\uparrow$'' means that higher values are better. The best record is highlighted in bold.
  }
\end{table}

%% file: table/cooperation.tex
\begin{table}
  \centering
  \setlength{\tabcolsep}{1mm}
\begin{tabular}{l|c|c|c}
\toprule
\multicolumn{1}{c|}{Method (CortexDebate)}    & DVC    & CVR  & CVR/DVC  \\ \midrule
without I and S factors     & 3.71   & 1.26   & 33.96     \\
with I and S factors      & 8.44  & 4.83  & 64.92         \\  \bottomrule
\end{tabular}
  \caption{\label{cooperation}
   Comparison results of our CortexDebate with and without considering intimacy (I) and self-orientation (S) factors.
  }
\end{table}

%% file: table/helpful.tex
\begin{table}
  \centering
  \setlength{\tabcolsep}{1.2mm}
\begin{tabular}{c|cc|cc}
\toprule
\multirow{2}{*}{Agent}               & \multicolumn{2}{c|}{FcG}   & \multicolumn{2}{c}{CortexDebate} \\ \cmidrule{2-5}
   & RA $\uparrow$  & Avg-DN $\downarrow$  & RA $\uparrow$  & Avg-DN $\downarrow$  \\   \midrule
Qwen     & 54.00   & 13.58    & 58.00  & 11.53       \\
Mistral    & 49.00  & 13.58  & 56.00   & 6.24      \\ 
Typhoon       & 47.00  & 13.58  & 53.00  & 8.83       \\
Llama & 51.00  & 13.58  & 56.00  & 11.37   \\
Gemma & 45.00  & 13.58  & 51.00  & 5.46   \\ \bottomrule
\end{tabular}
  \caption{\label{helpful}
   Task performance of every participated LLM agent under the fully-connected debating graph and our proposed CortexDebate. ``$\uparrow$'' means that higher values are better, and ``$\downarrow$'' means that lower values are better.
  }
\end{table}

%% file: conclusion.tex
\section{Conclusion}

In this paper, we propose a new MAD method termed ``CortexDebate'' to improve the reasoning abilities of multi-agent interaction systems. Specifically, our CortexDebate establishes a sparse debating graph among participating LLM agents, which reduces input information burdens of LLM agents. Besides, by integrating the McKinsey Trust Formula, our proposed MDM module conducts credible evaluations to gradually optimize the debating graph, making the debating process equal, in-depth, and effective. Due to the above designs, our method alleviates two major issues faced by existing MAD systems (\textit{i.e.}, too lengthy input contexts and overconfidence-caused unequal debates), and shows superior performance to various state-of-the-art MAD methods on various typical tasks. In the future, we plan to continue exploring the potential of CortexDebate in large-scale debates and complex tasks (\textit{i.e.}, domain expert systems).

%% file: limitations.tex
\section*{Limitations}
Despite the impressive performance of our proposed CortexDebate, we acknowledge that it has two main limitations. Firstly, as a multi-agent debate method, compared with single-agent methods, it is inevitable that there will be a decrease in efficiency and an increase in cost when solving tasks. Secondly, despite the success, the reasoning ability of LLM agents remains an important factor that limits the performance of CortexDebate. Although our proposed CortexDebate improves the debate strategy among LLM agents, mistakes may still occur due to the poor reasoning ability of LLM agents.

\section*{Acknowledgments}

This research is supported by NSF of China (Nos: 62336003, 12371510) and NSF of Jiangsu Province (No: BK20241469).

%% file: appendix/appendix_dataset_detail.tex
\section{Dataset Details}
\label{appendix_dataset_detail}

The eight datasets used in our experiments are classic datasets that are widely employed to evaluate the performance of agent-based methods. Here, we provide an introduction to the eight datasets used in our experiments.

\textbf{GSM-IC.}  It is a grade-school math problem dataset derived from GSM8K ~\citep{cobbe2021training}.  For each problem in GSM8K, GSM-IC keeps the base problem description and adds to it one irrelevant sentence that does not affect the solution of the problem.

\textbf{MATH.} It is a math dataset containing challenging competition mathematics problems. Each of them has a full step-by-step solution.

\textbf{MMLU.} It contains 57 types of multiple-choice problems, such as elementary mathematics, US history, computer science, and so on. To acquire high performance on the MMLU datasset, models must possess extensive world knowledge and strong problem-solving ability.

\textbf{MMLU-pro.} It contains questions sourced from multiple origins, such as MMLU, TheoremQA, and SciBench. Moreover, it expands the option number of each problem from 4 to 10.

\textbf{GPQA.} It contains 448 graduate-level question-answering problems, covering knowledge in various fields such as biology, physics, and chemistry.

\textbf{ARC-C.} It contains complex questions on natural science, presented in the form of multiple-choice options.

\renewcommand{\dblfloatpagefraction}{.9}
\input{table/table_appendix_experiments}
\input{table/appendix_table_additional_token}

\textbf{LongBench.} It is a dataset designed to evaluate the long-context understanding capabilities of models. It encompasses six major categories of tasks, including single-document QA, multi-document QA, summarization, few-shot learning, code completion, and synthetic tasks.

\textbf{SQuAD.} It is a dataset used to evaluate the reading comprehension ability of models. The dataset requires models to answer different questions from given long texts.

%% file: table/table_appendix_experiments.tex
\begin{table*}
  \centering
  \setlength{\tabcolsep}{3mm}
  \begin{tabular}{c|c|ccc|c}
\toprule
\multirow{2}{*}{Type}    & \multirow{2}{*}{Method}                      & MATH                 & MMLU-pro             & GPQA                 & LongBench                \\ \cmidrule{3-6}
&               & \multicolumn{3}{c|}{RA $\uparrow$}    & M-Avg $\uparrow$  \\ \midrule
No Debate            & MaV       & 47.40          & 46.30          & 29.10          & 43.35          \\ \midrule
\multirow{4}{*}{Full Debate}   & MLD        & 49.20          & 48.40          & 30.60          & 46.26          \\
& RECONCILE         & 50.70          & 53.10          & 30.80          & 48.33          \\
& ChatEval          & 49.90          & 49.30          & 31.10          & 51.23          \\
& PRD               & 51.20          & 54.20          & 32.40          & 47.67          \\  \midrule
\multirow{3}{*}{Part Debate}                       &GD                      & 50.30          & 51.30          & 34.20          & 54.58          \\
& ND     & 49.50          & 49.10          & 32.80          & 54.14          \\  \cmidrule{2-6}
& Ours   & \textbf{56.30} & \textbf{58.90} & \textbf{36.60} & \textbf{59.63} \\ \bottomrule
\end{tabular}
  \caption{\label{table_add_exp}
   Comparison results on the four datasets. The unit of all the results is ``$\%$''. ``$\uparrow$'' means that higher values are better. The best records under each metric are highlighted in bold.
  }
\end{table*}

%% file: table/appendix_table_additional_token.tex
\begin{table*}
  \centering
  \setlength{\tabcolsep}{3mm}
 \begin{tabular}{c|c|cccc}
\toprule
Type    &Method              & MATH            & MMLU-pro         & GPQA     & LongBench              \\ \midrule
No Debate   & MaV                 & \textcolor{gray}{1316.85}          & \textcolor{gray}{2268.10}          & \textcolor{gray}{2567.90}          & \textcolor{gray}{125034.39}           \\   \midrule
\multirow{4}{*}{Full Debate}   & MLD                 & 6408.39 & 11412.75 & 12868.37 & 585365.43          \\
& RECONCILE           & 6723.84 & 11334.12 & 14061.76 & 605688.14          \\
& ChatEval             & 5571.07 & 9219.88  & 12369.62 & 553114.66          \\
& PRD                 & 8849.70 & 14946.31 & 18851.29 & 815449.95          \\  \midrule
\multirow{3}{*}{Part Debate}   & GD          & 4217.23 & 7265.23  & 8817.46  & 447987.68          \\
& ND                  & 4175.27 & 6673.75  & 7971.60  & 394905.47          \\  \cmidrule{2-6}
& Ours & \textbf{3355.20} & \textbf{6001.33}  & \textbf{6503.76}  & \textbf{321109.57} \\ \bottomrule
\end{tabular}
  \caption{\label{table_appendix_additional_exp_token}
   Comparison results of average input context length on adopted datasets. Each result represents the average token number of input context. The results in gray indicate that they are not included in result comparison, since their corresponding method (MaV) is not MAD method. The best records among the MAD methods on each dataset are highlighted in bold.
  }
\end{table*}

%% file: appendix/supplementary_results.tex
\section{Supplementary Experimental Results}
\label{appendix_additional_results}

In this section, we provide the experimental result data involved in the charts which are presented in Sections~\ref{main_results} and ~\ref{performance_investigation}.

\renewcommand{\dblfloatpagefraction}{.9}
\input{table/append_result_token}
\input{table/round_accuracy_consensus}

For Figure~\ref{figure2}, we provide the data in Table~\ref{appendix_result_token}. Compared with the full debate methods (\textit{i.e.}, MLD, RECONCILE, ChatEval, and MPRC), our CortexDebate significantly reduces the length of the contextual input for each LLM agent, with a maximum reduction of 70.79\%. Moreover, compared with the part debate methods (\textit{i.e.}, GD and ND), our CortexDebate can reduce the input context length by at least 17.62\%.

For Figure~\ref{round_accuracy_consensus}, we provide the data in Table~\ref{appendix_round_acc_con}. After each debating round, our CortexDebate achieves the highest average score compared with the baseline methods, with a global maximal score of 69.41\% (67.30\% for the baseline methods).

%% file: table/append_result_token.tex
\begin{table*}
\small
  \centering
  \setlength{\tabcolsep}{1mm}
  \renewcommand{\arraystretch}{1.3}
 \begin{tabular}{c|c|cccccccc}
\toprule
Type    &Method              & GSM-IC           & MATH             & MMLU             & MMLU-pro         & GPQA             & ARC-C    & LongBench          & SQuAD             \\ \midrule
No Debate   & MaV                 & \textcolor{gray}{1161.18}          & \textcolor{gray}{1277.73}          & \textcolor{gray}{1670.48}          & \textcolor{gray}{2144.39}          & \textcolor{gray}{2653.92}          & \textcolor{gray}{1582.67}          & \textcolor{gray}{105020.51}          & \textcolor{gray}{4724.95}           \\   \midrule
\multirow{4}{*}{Full Debate}   & MLD                 & 6287.39          & 6397.97          & 7905.84          & 11213.07         & 12947.80         & 7605.01          & 525177.19          & 23185.66          \\
& RECONCILE           & 6196.09          & 6574.29          & 8409.61          & 11260.15         & 14107.64         & 8770.85          & 525765.69          & 24112.64          \\
& ChatEval            & 5600.22          & 5394.71          & 7325.20          & 9160.49          & 12252.56         & 6918.96          & 473070.19          & 20781.94          \\
& PRD                 & 7651.62          & 8652.23          & 11691.83         & 14829.46         & 18837.51         & 11232.72         & 735370.99          & 33350.02          \\  \midrule
\multirow{3}{*}{Part Debate}   & GD         & 3828.95          & 4139.04          & 6156.13          & 7159.45          & 8906.51          & 5381.72          & 367835.88          & 16073.50          \\
& ND                  & 3149.04          & 4207.18          & 4740.46          & 6647.27          & 8016.54          & 4311.26          & 314993.80          & 14218.49          \\  \cmidrule{2-10}
& Ours & \textbf{2413.35} & \textbf{3262.79} & \textbf{3727.65} & \textbf{5897.94} & \textbf{6340.26} & \textbf{3280.71} & \textbf{230956.05} & \textbf{11965.81} \\ \bottomrule
\end{tabular}
  \caption{\label{appendix_result_token}
   Comparison results of average input context length on eight datasets. Each result represents the average token number of input context. The results in gray indicate that they are not included in result comparison, since their corresponding method (MaV) is not MAD method. The best records among the MAD methods on each dataset are highlighted in bold.
  }
\end{table*}

%% file: table/round_accuracy_consensus.tex
\begin{table*}
\small
  \centering
\setlength{\tabcolsep}{6.5pt}
  \renewcommand{\arraystretch}{1.2}
\begin{tabular}{c|c|ccccc|ccccc}
\toprule
\multirow{2}{*}{Type}    & \multirow{2}{*}{Method} & \multicolumn{5}{c|}{Score (\%)}     & \multicolumn{5}{c}{Consensus (\%)}    \\ \cmidrule{3-12} 
                    &    & 1     & 2     & 3     & 4     & 5     & 1     & 2     & 3     & 4     & 5     \\ \midrule
\multirow{4}{*}{Full Debate}   & MLD                     & 55.99 & 58.76 & 60.32 & 59.63 & 59.82 & 52.88 & 66.88 & 82.00 & 87.88 & 86.50 \\
& RECONCILE               & 58.95 & 60.71 & 67.30 & 65.64 & 63.65 & 73.63 & 79.25 & 94.00 & 95.75 & 98.63 \\
& ChatEval                 & 58.13 & 63.86 & 61.74 & 63.01 & 62.32 & 55.75 & 72.13 & 67.63 & 78.63 & 88.50 \\
& PRD                      & 60.22 & 65.05 & 66.66 & 65.50 & 64.11 & 64.50 & 72.75 & 79.50 & 86.38 & 87.88 \\  \midrule
\multirow{3}{*}{Part Debate}   & GD                       & 56.94 & 59.04 & 64.19 & 64.87 & 63.91 & 48.13 & 54.75 & 74.00 & 78.88 & 81.75 \\
& ND                       & 56.21 & 58.65 & 57.52 & 61.65 & 62.19 & 43.38 & 56.63 & 66.13 & 73.00 & 77.75 \\  \cmidrule{2-12}
& Ours                    & 60.31 & 65.65 & 67.63 & 69.10 & 69.41 & 51.75 & 62.13 & 76.75 & 89.63 & 95.50 \\ \bottomrule
\end{tabular}
  \caption{\label{appendix_round_acc_con}
   Experimental results of average task scores and consensus proportion for all the methods after each debate round.
  }
\end{table*}

%% file: appendix/additional_experiment.tex
\section{Additional Experiments}
\label{appendix_additional_experiments}

In this section, we present large-scale experiments of our CortexDebate and baseline methods to further demonstrate the superiority of CortexDebate. 

\textbf{Experimental Setup.} For each task (\textit{i.e.}, math, world knowledge question answering, reasoning, and long-context understanding), we conduct experiments on the more challenging one of the two datasets (\textit{i.e.}, MATH, MMLU-pro, GPQA, and LongBench). For each adopted dataset, we experiment on a subset of 1000 examples. Besides, the backbone models and evaluation metrics used in the experiments are the same as mentioned in Section~\ref{main_experimental_setup}.

\textbf{Results.} The experimental results are presented in Table~\ref{table_add_exp}. Consistent with the experimental results reported in Table~\ref{main_result_acc}, our proposed CortexDebate achieves the best performance on all the datasets compared with baseline methods. For instance, CortexDebate achieves a maximal RA of 56.30\% on MATH dataset, 58.90\% on MMLU-pro dataset, 36.60\% on GPQA dataset, and 59.63\% on LongBench dataset, respectively. Moreover, for each method, we calculate the average token numbers of the contexts input to one LLM agent on each dataset and present the results in Table~\ref{table_appendix_additional_exp_token}. Compared with the full debate methods (\textit{i.e.}, MLD, RECONCILE, ChatEval, and MPRC), our CortexDebate significantly reduces the length of the contextual input for each LLM agent, with a maximum reduction of 65.50\%. Moreover, compared with the part debate methods (\textit{i.e.}, GD and ND), our CortexDebate can reduce the input context length by at least 17.40\%.

%% file: appendix/performance_investigation.tex
\section{Performance Investigation}
\label{pi}

\input{table/average_based}
\input{table/text_similarity}

In this section, we present additional in-depth investigations on our CortexDebate to further analyze its effectiveness.

\subsection{Average-based Edge Pruning}

As present in Equation~\eqref{11}, our CortexDebate uses the average value as the threshold to prune the edges. To evaluate the effectiveness of adopting the average-based threshold, we conduct experiments on MATH dataset. Specifically, we compare the strategy of debating with the agents above the average threshold (AAT) with three alternatives, namely Top-3 (debating with the top three agents), Bot-3 (excluding the bottom three agents), and AMT (debating with agents above the median threshold). As shown in Table~\ref{average_based}, AAT achieves superior performance than the others, which validates the effectiveness of the average-based edge pruning.

\subsection{Text Similarity Calculation}

In the experiments, we use the text-embedding-3-large model released by OpenAI as the embedding model. For calculation of text similarity, we use cosine similarity which is widely used due to its efficiency and robustness to varying input length, and strong empirical performance. To confirm this, on LongBench dataset, we compare our strategy with ag-nli-DeTS-sentence-similarity-v4 model (DeTS) that is trained on six natural language inference datasets and “DeepSeek R1 + prompt” (R1-P). The average task scores and variances over three runs are reported in Table~\ref{text_similarity}. We can find that our method outperforms the two baseline methods. Moreover, we observe that large language models tend to perform unstably when used for text similarity estimation, which is reflected by the higher variance across different runs.

\subsection{Confidence Recalibration}

As present in Equation~\eqref{recalibrate}, our CortexDebate adopts a recalibration strategy to mitigate overconfidence issue of LLM agents. Following the practice of~\cite{chen2023reconcile}, we conduct experiments on different combinations of thresholds. We present the average scores on eight adopted datasets in Table~\ref{confidence}. According to the results, this recalibration strategy leads to the improved performance, and the parameter setting in our paper leads to the best performance. Moreover, our method is generally insensitive to different threshold configurations. This demonstrates the effectiveness and robustness of the proposed confidence recalibration strategy.

\input{table/confidence}

%% file: table/average_based.tex
\begin{table}
  \centering
  \setlength{\tabcolsep}{5mm}
\begin{tabular}{cc}
\toprule
\multicolumn{1}{c}{Edge Pruning Strategy}     & RA $\uparrow$ \\ \midrule
Top-3     & 54.00         \\
Bot-3                & 56.00         \\
AMT       & 52.00         \\
AAT (Ours) & \textbf{57.00}     \\  \bottomrule
\end{tabular}
  \caption{\label{average_based}
   Task performance of CortexDebate with different edge pruning strategies. ``$\uparrow$'' means that higher values are better. The best record is highlighted in bold.
  }
\end{table}

%% file: table/text_similarity.tex
\begin{table}
  \centering
  \setlength{\tabcolsep}{3mm}
\begin{tabular}{cc}
\toprule
\multicolumn{1}{c}{Text Similarity Calculation}     & M-Avg $\uparrow$ \\ \midrule
DeTS     & 57.57{\tiny $\pm$0.69}         \\
R1-P                & 56.91{\tiny $\pm$3.10}         \\
Ours & \textbf{60.31{\tiny $\pm$0.32}}     \\  \bottomrule
\end{tabular}
  \caption{\label{text_similarity}
   Task performance of CortexDebate with different calculation strategies of text similarity. The format of the results is ``(average result)$\pm$(variance)''. ``$\uparrow$'' means that higher values are better. The best record is highlighted in bold.
  }
\end{table}

%% file: table/confidence.tex
\begin{table}
  \centering
  \setlength{\tabcolsep}{2mm}
\begin{tabular}{cc}
\toprule
Threshold Configurations     & Average Score $\uparrow$ \\ \midrule
w/o recalibration  & 68.62    \\
$[$0.9, 0.7, 0.2$]$    & 68.96    \\
$[$0.9, 0.6, 0.3$]$    & 69.03    \\
$[$0.8, 0.6, 0.2$]$    & 69.15    \\
$[$0.8, 0.5, 0.1$]$    & 68.84    \\
$[$0.7, 0.5, 0.2$]$    & 68.72    \\
$[$0.8, 0.6, 0.3$]$ (Ours)    & \textbf{69.41}    \\
  \bottomrule
\end{tabular}
  \caption{\label{confidence}
   Task performance of CortexDebate with different threshold configurations of the recalibration strategy. ``$\uparrow$'' means that higher values are better. The best record is highlighted in bold.
  }
\end{table}

%% file: appendix/evaluation_strategy.tex
\section{Introduction of Evaluation Strategies}
\label{appendix_evaluation_strategies}

Here we introduce the details of the evaluation strategies (\textit{i.e.}, self-evaluation, peer evaluation, part MDM) mentioned in Section~\ref{performance_investigation}.

\textbf{Self-evaluation.} In this strategy, each LLM agent is required to generate a confidence score for its generated answer. Each LLM agent will only debate with the LLM agents whose confidence scores are above the average of the entire graph.

\textbf{Peer Evaluation. } For each LLM agent, its answer is scored by other LLM agents, and the final score of the answer is the average of the received scores. Each LLM agent will only debate with the LLM agents whose scores are above the average of the entire graph.

\textbf{MDM (w/o $I_{d}$ and $S_{d}$ in Equation~\eqref{9}).} For McKinsey Trust Formula used in MDM module, this strategy only considers the first two aspects (\textit{i.e.}, credibility and reliability) which evaluate individual abilities, neglecting the last two aspects (\textit{i.e.}, intimacy and self-orientation) which evaluate the debate effectiveness between two LLM agents.

%% file: appendix/cortexdebate_prompt.tex
\section{Prompts in CortexDebate}
\label{appendix_prompt_cortexdebate}

\input{table/table_cortexdebate_prompt}

We provide the specific prompts of our proposed CortexDebate in Table~\ref{table_cor_prompt}. For initial answer generation, CortexDebate follows~\cite{kojima2022large} and prompts each LLM agent to solve the problem step by step. For answer regeneration, the prompt contains three parts: (a) An instruction that stimulates LLM agents to generate their new answers and self-confidence scores after scrutinizing other answers. (b) A description of the problem. (c) Some answers generated by other LLM agents.

%% file: table/table_cortexdebate_prompt.tex
\begin{table*}
  \centering
  \setlength{\tabcolsep}{3mm}
\begin{tabular}{cl}
\toprule
Type                      & \multicolumn{1}{c}{Prompt} \\ \midrule
\begin{tabular}[c]{@{}c@{}}Initial Answer\\ Generation\end{tabular} &  \begin{tabular}[c]{@{}p{0.6\textwidth}@{}} Question: \{the description of the question\} \\ Please think it step by step and generate an answer and an explanation for your answer. \\ Also, evaluate how confident you are that your answer is correct. Your confidence score should between 0 and 1. \\ The format of your answer must be:\\ \ \ \ \ \ \ \ \ \ \ Answer: (...)\\ \ \ \ \ \ \ \ \ \ \ Explanation: (...)\\ \ \ \ \ \ \ \ \ \ \ Confidence Score: (...)  \end{tabular}                         \\ \midrule
\begin{tabular}[c]{@{}c@{}}Answer\\ Regeneration\end{tabular}       &  \begin{tabular}[c]{@{}p{0.6\textwidth}@{}} Question: \{the description of the question\} \\ \\ There are some answers generated by other LLM agents: \\ One LLM agent answer: \{answer\} \\ One LLM agent answer: \{answer\} \\ ... ... \\Using these answers as additional information, please generate a new answer and an explanation for your answer. \\ Also, evaluate how confident you are that your answer is correct. Your confidence score should between 0 and 1. \\ The format of your answer must be:\\ \ \ \ \ \ \ \ \ \ \ Answer: (...)\\ \ \ \ \ \ \ \ \ \ \ Explanation: (...)\\ \ \ \ \ \ \ \ \ \ \ Confidence Score: (...) \end{tabular}                          \\ \bottomrule
\end{tabular}
  \caption{\label{table_cor_prompt}
   Prompts of our proposed CortexDebate used in the experiments.
  }
\end{table*}

%% file: appendix/appendix_baseline_introduction.tex
\section{Introduction of Baseline Methods}
\label{appendix_baseline_methods}

Here we introduce the details of the baseline methods (\textit{i.e.}, Multi-agent Voting, ChatEval, Multi-LLM Debate, RECONCILE, Peer Review Debate, GroupDebate, and Neighbor Debate) in our experiments.

\textbf{Multi-agent Voting.} This method adopts a majority voting strategy to aggregate responses from multiple LLM agents. Specifically, each LLM agent independently generates a response to the given question. The final prediction is then determined through majority voting.

\begin{figure*}[t]
\centering
  \includegraphics[width=.9\linewidth]{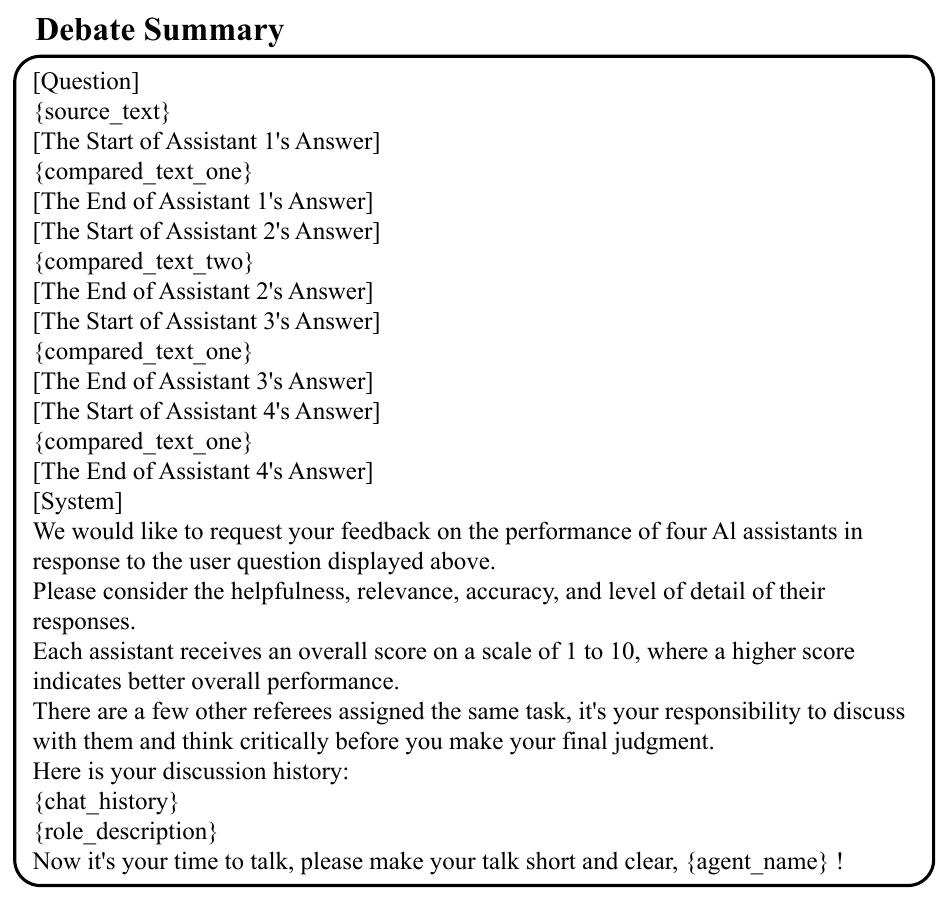}
  \caption {Debating summary prompt of ChatEval used in the experiments.}
  \label{ChatEval}
\end{figure*}

\textbf{ChatEval.} ChatEval uses an extra LLM agent to summarize the debating results in each round of debate. The specific prompt of debating summary used in the experiments is shown in Figure~\ref{ChatEval}. The summary text generated in the current round of debate will be input to each LLM agent as supplementary information in the next round of debate.

\begin{figure*}[t]
\centering
  \includegraphics[width=.9\linewidth]{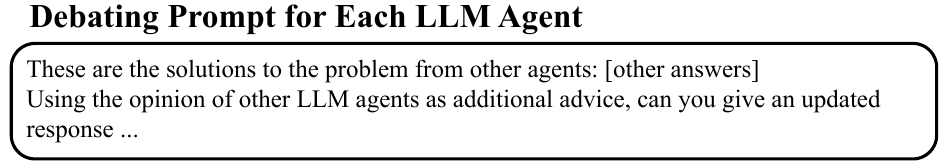}
  \caption {Prompt of Multi-LLM Debate used in the experiments.}
  \label{MLD}
\end{figure*}

\textbf{Multi-LLM Debate.} Firstly, each LLM generates an answer to the question. Then, each LLM reads and critiques the answers generated by other LLM agents, and generates its new answer. This step is repeated multiple times. After that, the final answer is obtained through majority voting among the answers generated by all the LLM agents in the last round of debate. The specific prompt used in the experiments is shown in Figure~\ref{MLD}.

\begin{figure*}[t]
\centering
  \includegraphics[width=.9\linewidth]{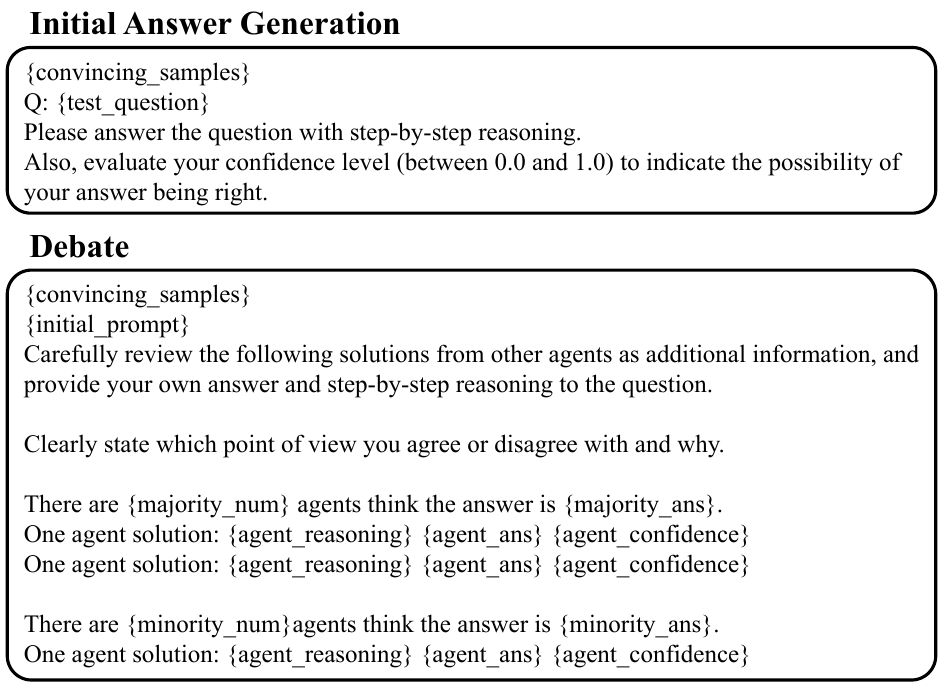}
  \caption {Prompt of RECONCILE used in the experiments.}
  \label{RECONCILE}
\end{figure*}

\textbf{RECONCILE.} Given a problem, each LLM first generates an answer and its uncertainty for the answer. Then
all LLM agents enter a multi-round debate. Each debating round consists of each LLM generating a revised answer and its new uncertainty based on the answers generated by all other LLM agents from the previous round. After the multi-round debate, RECONCILE obtains the final answer through majority voting. The specific prompt used in the experiments is shown in Figure~\ref{RECONCILE}.

\begin{figure*}[t]
\centering
  \includegraphics[width=.9\linewidth]{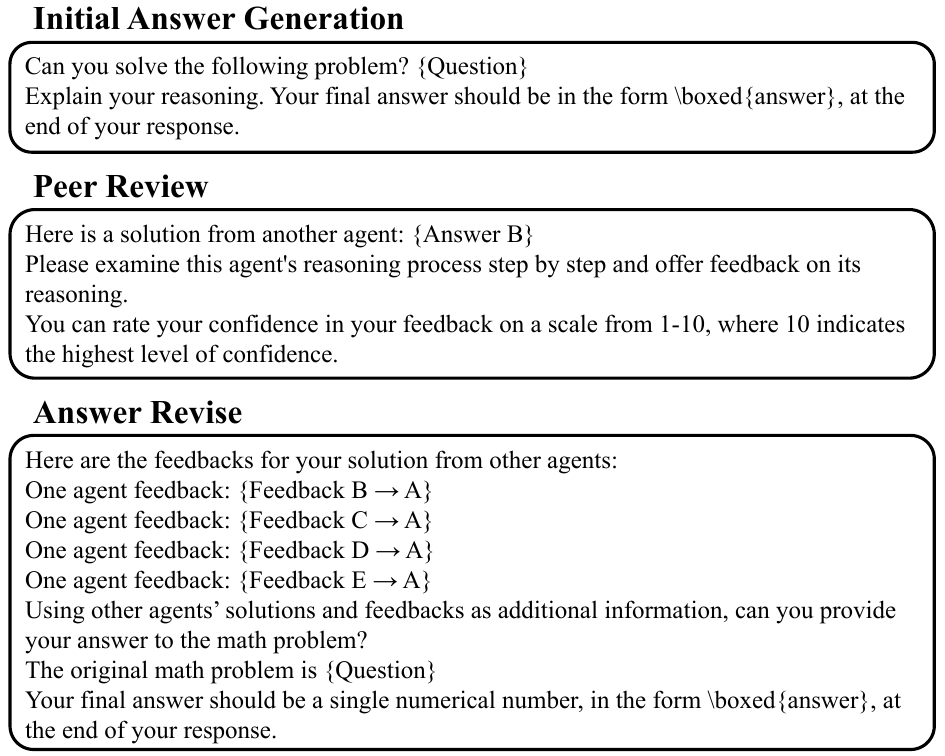}
  \caption {Prompt of Peer Review Debate used in the experiments.}
  \label{Peer_Review_Debate}
\end{figure*}

\begin{figure*}[t]
\centering
  \includegraphics[width=.9\linewidth]{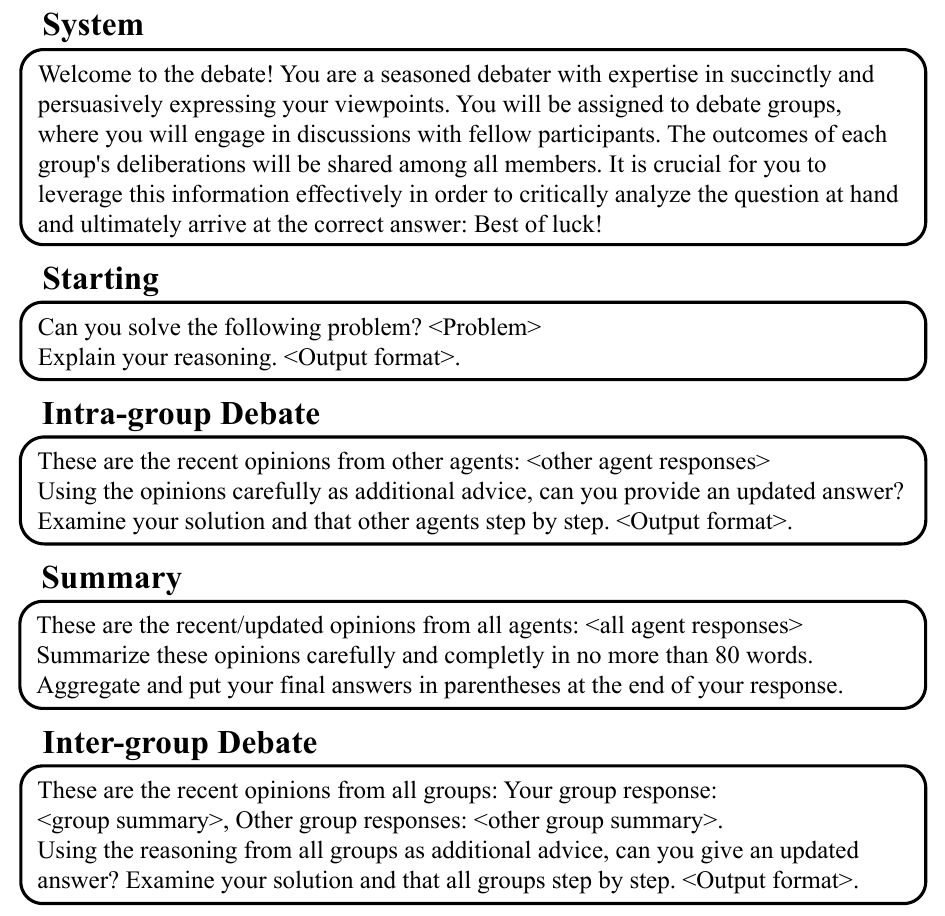}
  \caption {Prompt of GroupDebate Debate used in the experiments.}
  \label{GD}
\end{figure*}

\begin{figure*}[t]
\centering
  \includegraphics[width=.9\linewidth]{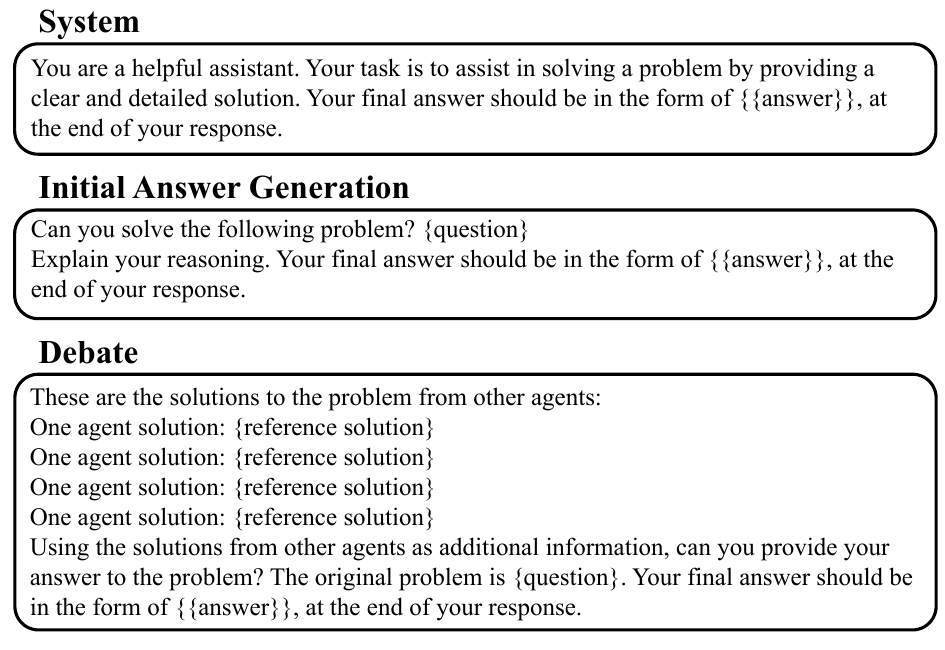}
  \caption {Prompt of Neighbor Debate used in the experiments.}
  \label{ND}
\end{figure*}

\textbf{Peer Review Debate.} Similar to RECONCILE, this method also evaluates all the answers in each round of debate. However, this method employs a peer review strategy where the answer generated by each LLM agent is evaluated by other LLM agents. The specific prompt used in the experiments is shown in Figure~\ref{Peer_Review_Debate}.

\textbf{GroupDebate.} This method divides the LLM agents into several debate groups, with each group conducting internal debates. After the internal debates, the result of each debate group is summarized and placed into a shared pool. After that, each group retrieves the debate summaries of all groups from the pool, which serve as the input for all the LLM agents in the next round. The specific prompt used in the experiments is shown in Figure~\ref{GD}.

\textbf{Neighbor Debate.} In this method, LLMs only debate with their neighbors. The specific prompt used in the experiments is shown in Figure~\ref{ND}.

%% file: appendix/algorithm.tex
\section{CortexDebate Algorithm}
\label{appendix_cortexdebate_algorithm}

In this section, we provide the detailed algorithm of our proposed CortexDebate. As present in Algorithm~\ref{alg:1}, we strictly follow Sections~\ref{main_preliminaries} and~\ref{main_method}, and provide the whole execution process of our proposed CortexDebate.

\begin{algorithm*}
\small
	\renewcommand{\algorithmicrequire}{\textbf{Input:}}
	\renewcommand{\algorithmicensure}{\textbf{Output:}}
	\caption{CortexDebate Method}
	\label{alg:1}
	\begin{algorithmic}[1]
		\REQUIRE Number of LLM agents $n$, set of LLM agents $\left \{A_{i} \right \}_{i=1}^{n}$, set of directed edges $\left \{E_{i\to j} \right \}_{i,j\in \left [1,2,...,n  \right ] }$, test question $Q$, maximum debating rounds $D$, answer extraction $ans\left ( \cdot  \right ) $
		\ENSURE Final answer $O_{final}$
            \FOR{$i=1$ to $n$}  
                \STATE $O_{i}^{0}$, $H_{i}^{0} \gets A_{i}\left (Q  \right )$  \textcolor{blue}{\COMMENT{Phase 1: Initial Answer Generation}}
                \STATE Recalibration $H_{i}^{0}$ based on Equation~\eqref{recalibrate}
                \STATE Calculate $C^{i}$ based on Equations~\eqref{3} and~\eqref{4}
                \STATE $P_{0}^{i}\gets 0$
            \ENDFOR
            \STATE $O \gets \left \{ O_{i}^{0} \right \}_{i=1}^{n} $  \textcolor{blue}{\COMMENT{Phase 2: Multi-round Debate}}
            \FOR{$d=1$ to $D$}
                \FOR{$i=1$ to $n$}
                    \STATE Calculate $R_{d}^{i}$ and $S_{d}^{i}$ based on Equations~\eqref{5} and~\eqref{8}, respectively
                    \FOR{$j=1$ to $n$}
                        \IF{$i\ne j$}
                            \STATE Calculate $\overline{{Sim}}_{d}$ and $I_{d}^{i}$ based on Equations~\eqref{6} and~\eqref{7}, respectively  
                            \STATE Calculate $W_{i\to j}^{d}$ based on Equation~\eqref{9}   \textcolor{blue}{\COMMENT{Phase 2, Step 1: Edge Weight Optimization}}
                        \ENDIF
                    \ENDFOR
                \ENDFOR
                \FOR{$j=1$ to $n$}
                    \STATE $Deb_{j}^{d}$, $Others_{j}^{d}\gets \emptyset$    \textcolor{blue}{\COMMENT{Phase 2, Step 2: Sparse Graph Establishment}}
                    \STATE Calculate $\overline{{W}}_{j}^{d}$ based on Equation~\eqref{10}
                    \FOR{$i=1$ to $n$}
                        \IF{$i\ne j$}
                            \STATE Calculate $W_{i\to j}^{d}$ based on Equation~\eqref{11}
                            \IF{$W_{i\to j}^{d}=1$}
                                \STATE $Deb_{j}^{d}\gets Deb_{j}^{d} \ \cup \ \left \{A_{i}  \right \} $
                                \STATE $Others_{j}^{d}\gets Others_{j}^{d} \ \cup \ \left \{O_{i}^{d-1}  \right \} $
                            \ENDIF
                        \ENDIF
                    \ENDFOR
                    \STATE $O_{i}^{d}$, $H_{i}^{d} \gets A_{j}\left (Q, \ Others_{j}^{d} \right )$  \textcolor{blue}{\COMMENT{Phase 2, Step 3: Answer Regeneration}}
                    \STATE Recalibration $H_{i}^{d}$ based on Equation~\eqref{recalibrate}
                \ENDFOR
                \STATE $is\_end \gets$ True    \textcolor{blue}{\COMMENT{Phase 2, Step 4: Debate Termination}}
                \FOR{$i=2$ to $n$}
                    \IF{$ans\left ( O_{1}^{d}  \right ) \ne ans\left ( O_{i}^{d}  \right ) $}
                        \STATE $is\_end \gets$ False
                        \STATE break
                    \ENDIF
                \ENDFOR
                \STATE $O \gets \left \{ O_{i}^{d} \right \}_{i=1}^{n} $
                \IF{$is\_end =$ True}
                    \STATE break
                \ENDIF
            \ENDFOR
            \STATE $o \gets$set$\left ( O_{1}, \ O_{2}, \ \cdots, \ O_{n} \right )$
            \STATE Get $O_{final}$ based on Equation~\eqref{14}   \textcolor{blue}{\COMMENT{Phase 3: Final Answer Generation}}
		\STATE \textbf{return} $O_{final}$
	\end{algorithmic}
\end{algorithm*}